\documentclass[conference]{IEEEtran}
\IEEEoverridecommandlockouts
\usepackage{cite}
\usepackage{amsmath,amssymb,amsfonts}
\usepackage{algorithmic}
\usepackage{graphicx}
\usepackage{textcomp}
\usepackage{xcolor}
\usepackage{epstopdf}
\usepackage{booktabs}
\usepackage{balance}
\usepackage{siunitx}
\usepackage{makecell}
\usepackage{tabularx}
\usepackage{multicol}
\usepackage{multirow}
\usepackage{booktabs}
\usepackage{graphicx}      

\usepackage{fancyhdr}
\fancypagestyle{mahmood}{
  \fancyhf{} 
  
  \fancyhead[C]{\footnotesize \textcopyright This work has been submitted to the IEEE for possible publication. Copyright may be transferred without notice, after which this version may no longer be accessible.}
}
\makeatletter
\let\ps@IEEEtitlepagestyle\ps@mahmood
\makeatother

\begin{document}
\title{On-Device Crack Segmentation for Edge Structural Health Monitoring\\}

\makeatletter
\newcommand{\linebreakand}{%
  \end{@IEEEauthorhalign}
  \hfill\mbox{}\par
  \mbox{}\hfill\begin{@IEEEauthorhalign}
}
\makeatother

\author{
\IEEEauthorblockN{Yuxuan Zhang, 
Ye Xu,  
Luciano Sebastian Martinez-Rau, 
Quynh Nguyen Phuong Vu,}

\IEEEauthorblockN{
Bengt Oelmann and 
Sebastian Bader}

\IEEEauthorblockA{
\textit{Department of Computer and Electrical Engineering, Mid Sweden University, Sundsvall, Sweden} \\
yuxuan.zhang@miun.se}

}
 
\maketitle

\begin{abstract}
Crack segmentation can play a critical role in Structural Health Monitoring (SHM) by enabling accurate identification of crack size and location, which allows to monitor structural damages over time. However, deploying deep learning models for crack segmentation on resource-constrained microcontrollers presents significant challenges due to limited memory, computational power, and energy resources. To address these challenges, this study explores lightweight U-Net architectures tailored for TinyML applications, focusing on three optimization strategies: filter number reduction, network depth reduction, and the use of Depthwise Separable Convolutions (DWConv2D). Our results demonstrate that reducing convolution kernels and network depth significantly reduces RAM and Flash requirement, and inference times, albeit with some accuracy trade-offs. Specifically, by reducing the filer number to 25\%, the network depth to four blocks, and utilizing depthwise convolutions, a good compromise between segmentation performance and resource consumption is achieved. This makes the network particularly suitable for low-power TinyML applications. This study not only advances TinyML-based crack segmentation but also provides the possibility for energy-autonomous edge SHM systems. 
\end{abstract}



\section{Introduction}
In modern civil engineering, aerospace, and other large-scale infrastructures, structural safety directly impacts public welfare and economic benefits \cite{10273776}. Structural health monitoring (SHM) enables timely performance tracking and failure prevention \cite{zhangtim2023a, 9964134}. Since cracks are critical early indicators of localized stress or fatigue that may lead to irreversible damage \cite{10299555}, their precise detection and characterization are fundamental for safety assessments and extending service life. Unlike traditional manual inspection, crack segmentation provides an automated and quantitative approach by delineating crack boundaries. This enables accurate measurement of crack width, length, and propagation, which are key to assessing structural integrity and guiding maintenance decisions \cite{10286091}. 

In practical applications—such as bridges or hydraulic facilities—limited power supply and communication infrastructures hinder the continuous operation of power-hungry devices. Moreover, comprehensive real-time monitoring requires numerous sensor nodes, and equipping each with high-performance hardware quickly escalates costs and maintenance burdens. These challenges are even more pronounced for energy-autonomous systems with strict power budgets \cite{10572267}, making high-accuracy crack detection under resource constraints a key issue.

Deep learning (DL) has made significant strides in image segmentation \cite{10416966}. Architectures like U-Net and its variants effectively capture detailed boundary information for tasks including crack detection. However, state-of-the-art models such as PSPNet (21.07M parameters \cite{8100143}), DDRNet (20.18M \cite{9996293}), DeeplabV3+ (12.38M \cite{chen2017rethinking}), and RUCNet (25.47M \cite{s23010053}) require substantial storage and computation. Although they achieve high segmentation performance (mIoU of 0.7–0.8) on servers or GPUs, their heavy resource demands make real-time inference on edge devices challenging, highlighting the need for lightweight adaptations.

Tiny Machine Learning (TinyML) has recently garnered increasing attention \cite{10329945,adin2023b,huang2024b,Ud2024} for addressing these issues by deploying tailored models on low-power microcontrollers (MCUs). TinyML significantly reduces memory usage and inference power consumption while largely preserving model accuracy \cite{zhang2025survey}. For example, Zhang et al. \cite{zhangsas2024b} compared lightweight convolutional neural network models for crack classification under the TinyML framework, Chen et al. proposed a low-power on-device predictive maintenance system based on self-powered sensing and TinyML \cite{10229236}. Nonetheless, crack segmentation using TinyML remains unexplored.

To achieve on-device crack segmentation for energy-autonomous SHM at the edge, this paper systematically compares three lightweight strategies based on the U-Net architecture: (i) decreasing the number of convolutional filters to reduce parameters and inference overhead; (ii) reducing network depth to lower model size and computational complexity; and (iii) replacing standard convolutions with sparse convolutions to eliminate redundant computations while retaining sensitivity to crack details. By integrating these strategies with TinyML, we evaluate their performance on a low-power, resource-constrained MCU in terms of model accuracy, memory requirements, inference time, and energy consumption, and discuss their feasibility and limitations for energy-autonomous sensor nodes. This comprehensive analysis not only offers specialized solutions for crack segmentation on low-power devices but also serves as a valuable reference for deploying lightweight DL models in edge SHM systems.

\section{Dataset}
In this study, we constructed a comprehensive crack segmentation dataset by integrating images and masks from multiple publicly available datasets, including DeepCrack \cite{liu2019deepcrack}, Crack500 \cite{zhang2016road}, GAPs384 \cite{eisenbach2017how}, CrackTree \cite{zou2012cracktree}, CFD \cite{shi2016automatic}, and AEL \cite{7572082}. This dataset includes a total of 5000 samples (images and corresponding masks), effectively covering a wide range of crack patterns and scene variations. The dataset was split into 3500 training samples, 750 validation samples, and 750 test samples, following a 70\%, 15\% and 15\% ratio. The examples of crack images and masks are present in Fig. \ref{fig:datasetsamples}.

\begin{figure}[t]
\centerline{\includegraphics[width = 0.45\textwidth]{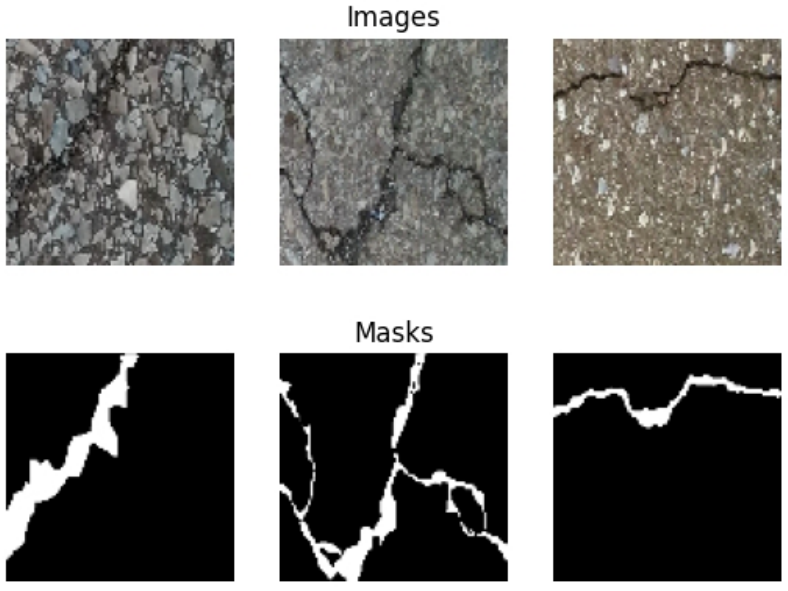}}

\caption{Image and mask samples of the dataset used in this study.}
\label{fig:datasetsamples}
\end{figure}

\section{Methods}

In this section, the method of the lightweight design and implementation process based on U-Net as our baseline is provided. The structure of the baseline U-net model is shown in Table \ref{tab:unet_structure}. The input image size is set to 96x96x3 to simultaneously avoid excessive information loss and minimize computational costs. In Subsection \ref{subSecLightweight}, we introduce three strategies for model size reduction, including reducing the number of convolution filters, reducing network depth, and replacing convolutional layers with sparse convolutional layers. In Subsection \ref{subSecModelTrain}, we detail the data preprocessing procedures, the loss function, training hyperparameters, and performance evaluation metrics as well as the experimental environment used in this study. Finally, Subsection \ref{subSecTinyML} will present how the lightweight models are deployed on resource-constrained hardware using a TinyML toolchain, including the development board employed and the model quantization strategy adopted.

\begin{table}[htbp]
\caption{Baseline U-Net Architecture}
\label{tab:unet_structure}
\centering
\resizebox{\linewidth}{!}{%
\begin{tabular}{@{}llc@{}}
\toprule
\textbf{Parameters} & \multicolumn{2}{c}{31,031,745}\\
\midrule
\textbf{Stage} & \textbf{Layers / Operations} & \textbf{Filters} \\ 
\midrule
\textbf{Input} & Input image & $(96 \times 96 \times 3)$ \\ 
\midrule
\textbf{Down 1} & Conv2D(3×3) ×2, ReLU, MaxPool(2×2) & 64 \\
\textbf{Down 2} & Conv2D(3×3) ×2, ReLU, MaxPool(2×2) & 128 \\
\textbf{Down 3} & Conv2D(3×3) ×2, ReLU, MaxPool(2×2) & 256 \\
\textbf{Down 4} & Conv2D(3×3) ×2, ReLU, MaxPool(2×2) & 512 \\
\midrule
\textbf{Bottleneck} & Conv2D(3×3) ×2, ReLU & 1024 \\
\midrule
\textbf{Up 1} & 
Conv2DTrans(2×2), Concat (Down 4) & 512 \\
& Conv2D(3×3) ×2, ReLU & \\
\textbf{Up 2} & 
Conv2DTrans(2×2), Concat (Down 3) & 256 \\
& Conv2D(3×3) ×2, ReLU & \\
\textbf{Up 3} & 
Conv2DTrans(2×2), Concat (Down 2) & 128 \\
& Conv2D(3×3) ×2, ReLU & \\
\textbf{Up 4} & 
Conv2DTrans(2×2), Concat (Down 1) & 64 \\
& Conv2D(3×3) ×2, ReLU & \\
\midrule
\textbf{Output} & Conv2D(1×1), Sigmoid & 1 \\
\bottomrule
\end{tabular}
}
\end{table}

\subsection{Model Size Reduction Strategies}
\label{subSecLightweight}

\textbf{Reducing the Number of Convolution Filters:} In this strategy, we systematically reduce the number of convolution filters at each layer of the U-Net architecture while preserving its standard encoder–decoder design. The baseline U-Net doubles the number of filters at each downsampling step, starting from 64 and increasing up to 1024 at the bottleneck layer, followed by symmetrical upsampling and concatenation layers that ultimately reduce the filters to 64 before the final output layer (Conv2D 1×1). In this study, we uniformly reduce the number of filters by half, quarter, one-eighth, and one-sixteenth of the original counts. This effectively decreases the network's parameter count and computational cost.

\textbf{Reducing Network Depth:} Another effective strategy for reducing the U-Net model size is to reduce its network depth by decreasing the number of downsampling and upsampling layers. In the baseline U-Net, the encoder path consists of five convolutional blocks, each followed by a max-pooling operation that reduces the spatial resolution while doubling the number of filters. This results in five corresponding upsampling blocks in the decoder path, creating a symmetric U-shaped structure. By systematically reducing the depth of U-Net from five layers to four or even three layers, the number of parameters and computations required for inference will be significantly reduced.

\textbf{Replacing Standard Convolutions with Sparse Convolutions:} The last model size reduction strategy for U-Net is to replace standard 3×3 convolutions with Depthwise Separable Convolutions, which decompose each convolution into a Depthwise Convolution and a Pointwise 1×1 Convolution proposed in \cite{howard2017mobilenets}. This reduces parameters and FLOPs while maintaining the encoder–decoder structure and receptive field. However, it may limit the model's ability to learn complex features, possibly affecting segmentation accuracy. We evaluate this trade-off by comparing U-Net variants using standard and sparse convolutions under identical conditions.

\subsection{Model Training and Validation}
\label{subSecModelTrain}

The models were trained with a Windows 10 64-bit operating system, an Intel® Core i9 12900 CPU, 32GB RAM, and a single RTX 3090 GPU with 24GB memory. We used the TensorFlow 2.8.0 framework, and all models were trained with a learning rate of 0.0001 and batch size of 8 over 15 epochs.

The loss function used in this study is Focal Tversky Loss which is defined in (\ref{lossfunction}). Herein, \( TP \), \( FP \), and \( FN \) denote the number of true positives, false positives, and false negatives, respectively. The parameters \( \alpha \) and \( \beta \) control the trade-off between false positives and false negatives, allowing the loss to be more sensitive to specific types of misclassifications. The focusing parameter \( \gamma \) adjusts the importance of easy versus hard examples, effectively emphasizing challenging samples during training. A small constant \( \epsilon \) is added to avoid division by zero. In this study, based on \cite{NGUYEN2023116988}, we set \( \alpha = 0.3 \), \( \beta = 0.7 \), \( \gamma = \frac{4}{3} \), and \( \epsilon = 10^{-6} \).

\begin{equation}
\label{lossfunction}
\mathcal{L}_{FTL} = \left( 1 - \frac{TP + \epsilon}{TP + \alpha \cdot FP + \beta \cdot FN + \epsilon} \right)^\gamma
\end{equation}

For the evaluation of the models' segmentation performance, the F1-score is used as a key metric as it balances precision and recall in a single value. It ensures that the model's ability to accurately identify true positives is not masked by the number of false positives and negatives. The F1-score is calculated as the harmonic mean of the precision and recall as given in (\ref{equF1}). The precision and recall are given in (\ref{equPreRecall}). 

\begin{equation}
\label{equF1}
    F1 = \frac{2 * Precision * Recall}{Precision + Recall}
\end{equation}
\begin{equation}
\label{equPreRecall}
    Precision = \frac{TP}{TP + FP}, 
    Recall = \frac{TP}{TP + FN}
\end{equation}

Also, mean Intersection over Union (mIoU) defined in (\ref{equmIoU}) is used in this study for semantic segmentation tasks. In (\ref{equmIoU}), \( C \) is the total number of classes, and \( \mathrm{TP}_c \), \( \mathrm{FP}_c \), and \( \mathrm{FN}_c \) represent the true positives, false positives, and false negatives for class \( c \), respectively. mIoU is particularly important in semantic segmentation, because it balances precision and recall by penalizing both false positives and false negatives. This makes it more robust to class imbalance compared to accuracy-based metrics. Additionally, mIoU provides a per-class evaluation, ensuring that performance on minority classes is not overshadowed by the majority class. Consequently, it serves as a reliable indicator of a model's overall segmentation capability, especially in scenarios with diverse object categories and complex scene compositions.

\begin{equation}
\label{equmIoU}
    \mathrm{mIoU} = \frac{1}{C} \sum_{c=1}^{C} \frac{\mathrm{TP}_c}{\mathrm{TP}_c + \mathrm{FP}_c + \mathrm{FN}_c}.
\end{equation}

\subsection{Tiny Machine Learning}
\label{subSecTinyML}


\begin{table}[t]
\caption{Technical specifications of OpenMV Cam H7 Plus}
\begin{center}
\begin{tabular}{cc}
\toprule
\textbf{\textit{Parameter}}& \textbf{\textit{OpenMV Cam H7 Plus Development Board}} \\
\midrule
MCU                 &  STM32H743II  \\
CPU Core            &  ARM Cortex M7 \\
CPU Frequency       &  480MHz   \\
RAM                 &  32 MB SDRAM + 1 MB SRAM    \\
Flash               &  32 MB external flash + 2 MB internal flash   \\
Voltage             &  3.3V         \\
\bottomrule
\end{tabular}
\label{tab:BoardDetails}
\end{center}
\end{table}

To deploy the lightweight U-Net models on resource-constrained devices, we utilized a TinyML workflow. In this study, we used the OpenMV Cam H7 Plus as the target development board. The board is powered by an STM32H743II ARM Cortex-M7 processor and integrated with an OV5640 image sensor, with detailed specifications shown in Table \ref{tab:BoardDetails}.
Although the target development board has 32 MB of SDRAM, the maximum deep learning model usage it can support is 4 MB according to the current firmware version (v4.5.6).
The end-to-end workflow begins with model training in TensorFlow, followed by conversion to TensorFlow Lite (TFLite) format. To optimize the model for embedded deployment, we performed post-training quantization, converting the model weights from float32 to int8. This quantization step significantly reduces the model size and computational requirements. The quantized TFLite model is then deployed on the OpenMV Cam H7 Plus using MicroPython and TensorFlow Lite for Microcontrollers (TFLM)\footnote{https://ai.google.dev/edge/litert}. This TinyML workflow is developed and deployed using the OpenMV IDE (version 4.4.7)\footnote{https://openmv.io/}. It is noteworthy that the OpenMV Cam H7 Plus captures images at 320×240 resolution using the integrated OV5640 sensor. To align the input dimensions with the crack segmentation model, the 96×96 central region is cropped from each frame and used as the model input.

\section{Results and Discussion}

\begin{figure}[t]
\centerline{\includegraphics[width = 0.35\textwidth]{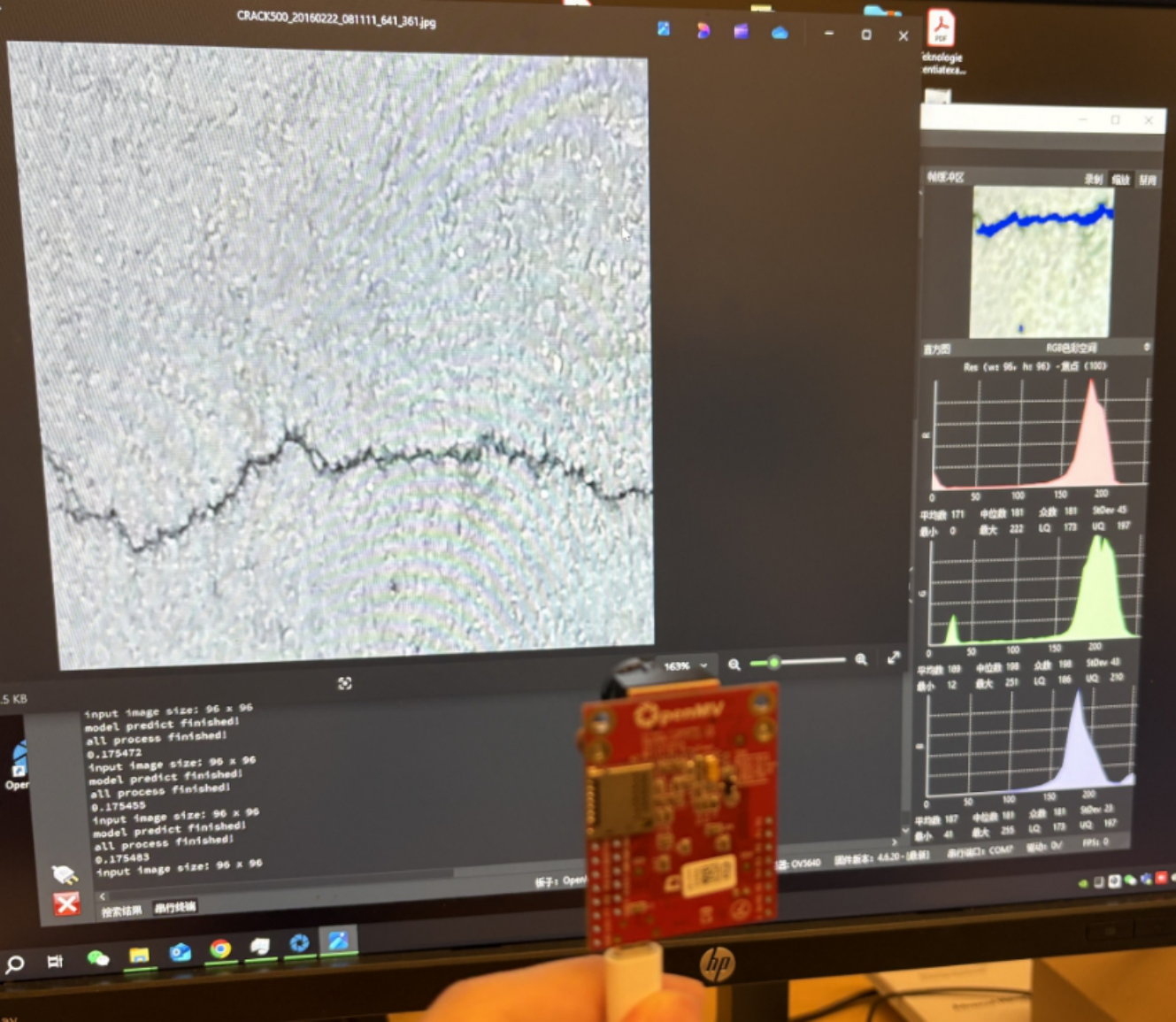}}
\caption{On-device real-time crack segmentation application.}
\label{fig:CrackSegApp}
\end{figure}

\begin{table*}[htbp]
\centering
\caption{On-device Results of Lightweighting Strategies and TinyML Deployment \\ 
 (Papameters in Thousand, RAM and flash in Kilobyte, time in Second, single segmentation energy in Joule)}
\renewcommand{\arraystretch}{1.2} 
\resizebox{\textwidth}{!}{
\begin{tabular}{l c | c c | c c | c c c c | c c c c}
\hline
\multirow{2}{*}{\textbf{Filters}} 
& \multirow{2}{*}{\textbf{Params.}} 
& \multicolumn{2}{c|}{\textbf{Float 32}} 
& \multicolumn{2}{c|}{\textbf{Int 8}} 
& \multicolumn{4}{c|}{\textbf{Model Inference}} 
& \multicolumn{4}{c}{\textbf{Whole System (including camera)}} \\
\cline{3-14}
& 
& \textbf{F1-score} & \textbf{mIoU}
& \textbf{F1-score} & \textbf{mIoU}
& \textbf{RAM} & \textbf{Flash} & \textbf{Time} & \textbf{Energy}
& \textbf{RAM} & \textbf{Flash} & \textbf{Time} & \textbf{Energy}\\
\hline
\multicolumn{14}{c}{\textbf{U-Net 5 ConvBlocks Conv2D}} \\
x1
& 31,031.75 
& 0.685 & 0.682 
& 0.618 & 0.658 
& 3,807.9 & 30,522.7 & n/a & n/a
& n/a & n/a & n/a &n/a\\
x1/2 
& 7,760.10
& 0.705 & 0.693
& 0.673 & 0.683
& 1,909.3 & 7,687.4 & 13.52 & 18.252
& 3,345.8 & 7,690.0 & 13.54 & 18.279\\
x1/4 
& 1,941.11
& 0.691 & 0.689
& 0.643 & 0.672
& 959.5 & 1,957.9 & 2.95 & 3.9825
& 3,343.7 & 1,960.9 & 2.97 & 4.0095\\
x1/8 
& 485.82
& 0.659 & 0.672
& 0.011 & 0.483
& 484.5 & 515.0 & 0.74 & 0.999
& 1,057.6 & 516.5 & 0.77 & 1.0395\\
x1/16 
& 121.73
& 0.588 & 0.640
& 0.552 & 0.635
& 247.5 & 147.7 & 0.22 & 0.297
& 675.6 & 149.5 & 0.24 & 0.324\\
\hline
\multicolumn{14}{c}{\textbf{U-Net 5 ConvBlocks DWConv2D}} \\
x1 
& 4,222.69
& 0.677 & 0.683
& 0.674 & 0.681
& 3,812.5 & 4,323.9 & n/a & n/a
& n/a & n/a & n/a  &n/a\\
x1/2 
& 1,066.88
& 0.647 & 0.668
& 0.635 & 0.663
& 1,912.1 & 1,153.5 & 5.42 & 7.317
& 3,347.4 & 1,155.5 & 5.45 & 7.3575\\
x1/4 
& 272.34
& 0.636 & 0.665
& 0.629 & 0.662
& 962.2 & 333.8 & 1.51 & 2.0385
& 1,822.1 & 336.1 & 1.53 & 2.0655\\
x1/8 
& 70.90
& 0.649 & 0.672
& 0.651 & 0.674
& 487.5 & 115.4 & 0.47 & 0.6345
& 1058.7 & 115.8 & 0.50 & 0.675\\
x1/16 
& 19.15
& 0.597 & 0.653
& 0.603 & 0.657
& 250.5 & 54.0 & 0.17 & 0.2295
& 677.8 & 56.4 & 0.19 & 0.2565\\
\hline
\multicolumn{14}{c}{\textbf{U-Net 4 ConvBlocks Conv2D}} \\
x1 
& 7,697.35
& 0.705 & 0.692
& 0.707 & 0.694
& 3,783.8 & 7,616.4 & n/a & n/a
& n/a & n/a & n/a &n/a\\
x1/2 
& 1,925.60
& 0.693 & 0.687
& 0.695 & 0.688
& 1,893.4 & 1,938.1 & 9.36 & 12.636
& 3,329.5 & 1,939.4 & 9.39  & 12.6765\\
x1/4 
& 482.03
& 0.643 & 0.664
& 0.645 & 0.665
& 951.5 & 506.0 & 2.10 & 2.835
& 1,812.4 & 508.6 & 2.13 & 2.8755\\
x1/8 
& 120.83
& 0.639 & 0.664
& 0.596 & 0.653
& 479.4 & 142.6 & 0.54 & 0.729
& 1,052.2 & 144.9 & 0.56  & 0.756\\
x1/16 
& 30.37
& 0.576 & 0.636
& 0.531 & 0.617
& 243.2 & 48.8 & 0.20 & 0.27
& 672.6 & 51.2 & 0.23 & 0.3105\\
\hline
\multicolumn{14}{c}{\textbf{U-Net 4 ConvBlocks DWConv2D}} \\
x1 
& 1,053.41
& 0.669 & 0.680
& 0.675 & 0.683
& 3,781.3 & 1,131.4 & n/a & n/a
& n/a & n/a & n/a &n/a\\
x1/2
& 268.67
& 0.650 & 0.670
& 0.654 & 0.672
& 1,896.6 & 322.9 & 4.23 & 5.7105
& 3,331.5 & 325.7 & 4.24 & 5.724\\
x1/4 
& 69.84
& 0.650 & 0.675
& 0.652 & 0.676
& 952.5 & 108.0 & 1.23 & 1.6605
& 1,813.2 & 110.1 & 1.26 & 1.701\\
x1/8 
& 18.81
& 0.633 & 0.667
& 0.645 & 0.674
& 481.3 & 47.9 & 0.40 & 0.54
& 1,054.5 & 50.3 & 0.42  & 0.567\\
x1/16 
& 5.39
& 0.566 & 0.640
& 0.580 & 0.649
& 246.0 & 29.7 & 0.16 & 0.216
& 674.3 & 32.1 & 0.18 & 0.243\\
\hline
\multicolumn{14}{c}{\textbf{U-Net 3 ConvBlocks Conv2D}} \\
x1 
& 1,862.85
& 0.699 & 0.691
& 0.701 & 0.692
& 3,762.5 & 1,868.4 & n/a & n/a
& n/a & n/a & n/a &n/a\\
x1/2 
& 466.53
& 0.661 & 0.673
& 0.663 & 0.675
& 1,885.4 & 485.6 & 4.94 & 6.669
& 3,320.4 & 487.1 & 4.97 & 6.7095\\
x1/4 
& 117.04
& 0.630 & 0.662
& 0.635 & 0.665
& 945.0 & 134.3 & 1.29 & 1.7415
& 1,806.8 & 136.5 & 1.31 & 1.7685\\
x1/8 
& 29.47
& 0.569 & 0.637
& 0.540 & 0.626
& 475.0 & 43.8 & 0.40 & 0.54
& 1,049.3 & 46.1 & 0.43 & 0.5805\\
x1/16 
& 7.47
& 0.484 & 0.600
& 0.471 & 0.605
& 240.2 & 19.9 & 0.17 & 0.2295
& 673.3 & 22.2 & 0.20 & 0.27\\
\hline
\multicolumn{14}{c}{\textbf{U-Net 3 ConvBlocks DWConv2D}} \\
x1 
& 255.20
& 0.576 & 0.637
& 0.582 & 0.640
& 3,763.3 & 299.7 & n/a & n/a
& n/a & n/a & n/a &n/a\\
x1/2
& 66.18
& 0.595 & 0.648
& 0.597 & 0.650
& 1,886.3 & 96.8 & 3.00 & 4.05
& 3,321.5 & 99.0 & 3.02 & 4.077\\
x1/4 
& 17.74
& 0.585 & 0.647
& 0.590 & 0.650
& 946.1 & 40.5 & 0.92 & 1.242
& 1,808.7 & 42.8 & 0.95 & 1.2825\\
x1/8 
& 5.05
& 0.574 & 0.647
& 0.567 & 0.643
& 477.0 & 23.6 & 0.32 & 0.432
& 1,050.4 & 26.0 & 0.35 & 0.4725\\
x1/16 
& 1.58
& 0.499 & 0.613
& 0.499 & 0.612
& 242.5 & 18.0 & 0.13 & 0.1755
& 671.5 & 20.3 & 0.16 & 0.216\\
\hline
\end{tabular}
}
\label{table:bigcomparision}
\end{table*}

This section presents the results of different reduction strategies on the target device, focusing on parameter count, F1-score, and mIoU for model performance, and RAM usage, flash usage, inference time, and energy consumption at the device level. Both standalone model inference and the complete system (including image capture and preprocessing) are evaluated. Real-world crack segmentation is shown in Fig. \ref{fig:CrackSegApp}, and detailed results are summarized in Table \ref{table:bigcomparision}. In comparison to the overall system consumption, deploying U-Net variant models in isolation consumes 35–60\% of the RAM, while flash memory usage, inference time, and energy consumption occupy for nearly 99\% of the total system overhead, underscoring the necessity of model optimization. Notably, x1 versions of U-Net models exceed the 4 MB RAM limit of the target board, making them undeployable. The following sections analyze the impact of reducing convolutional filters, decreasing network depth, and using sparse convolutions on model performance and inference time.

\subsection{Impact of Filter Number Reduction}
This section examines the impact of proportionally reducing the number of convolution kernels on segmentation performance, as well as on RAM, Flash, and inference time consumption (see Fig. \ref{fig:filters}). The presented averages are computed across various convolution types (Conv2D and DWConv2D) and model depths (5, 4, and 3 blocks) under different filter number reduction (x1, x1/2, x1/4, x1/8, x1/16).

Results show that decreasing the number of kernels leads to declines in both F1-score and mIoU for Float32 and Int8 quantization modes. For the Float32 model, reducing the kernel count from x1 to x1/16 lowers the F1-score from 0.6685 to 0.5517 and mIoU from 0.6773 to 0.6302, indicating a significant deterioration in segmentation performance. In contrast, the Int8 model exhibits a more pronounced performance drop, particularly at x1/8 and x1/16, suggesting that quantization errors are amplified in low-parameter configurations. In terms of resource consumption, RAM usage is reduced from 3873.8 KB to 251.0 KB and Flash usage from 7805.8 KB to 54.2 KB when the kernel count decreases from x1 to x1/16. Inference time is similarly reduced, from 6.74 s (x1/2) to 0.17 s (x1/16). 

\begin{figure}
    \centering
    \includegraphics[width=\linewidth]{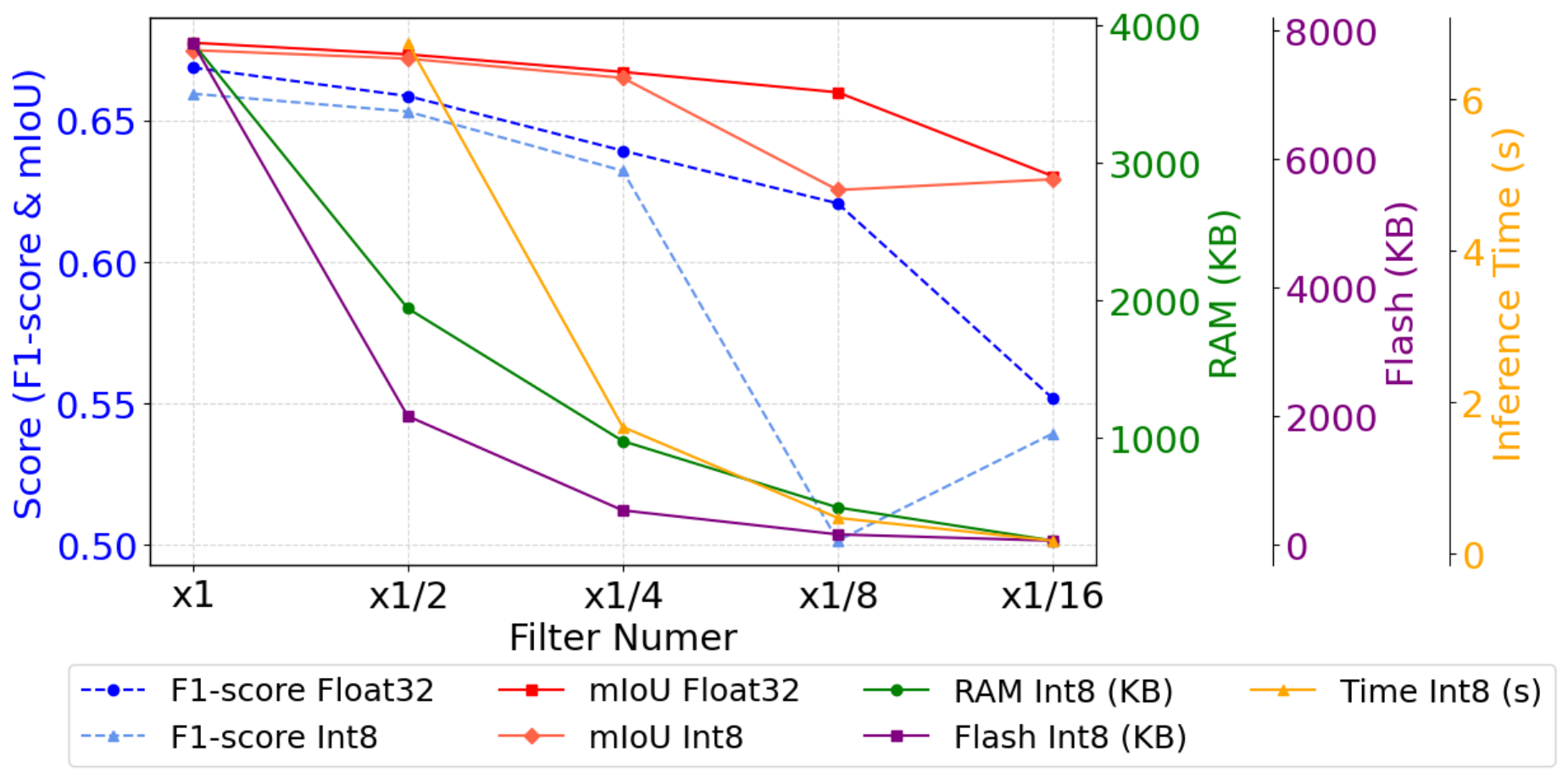}
    \caption{Average F1-score, mIoU, RAM, flash, and inference time vs filter number reduction}
    \label{fig:filters}
\end{figure}

\subsection{Impact of Model Depth Reduction}

This section investigates the impact of reducing network depth (from 5 to 3 blocks) on the segmentation performance and resource consumption of U-Net models across various filter number reduction (x1, x1/2, x1/4, x1/8, x1/16) and convolution types (Conv2D, DWConv2D). Figure \ref{fig:modeldepth} illustrates the trend computed as the average over all configurations, with network depth as the only variable.

Under Float32 quantization, shallower networks incur a performance drop: the F1-score decreases from 0.6534 (5 Blocks) to 0.5872 (3 Blocks), while the mIoU declines from 0.6718 to 0.6454. In contrast, the Int8 quantized model exhibits a non-monotonic trend; the F1-score increases from 0.5688 (5 Blocks) to 0.6381 (4 Blocks) before decreasing to 0.5842 (3 Blocks), and the mIoU remains relatively stable across depths. In terms of resource metrics, reducing network depth brings significant improvements. RAM consumption is slightly reduced (from 1518.4 KB to 1497.1 KB), Flash usage is substantially lowered (from 4790.8 KB to 310.1 KB), and inference time is significantly shortened (from 3.12 s to 1.39 s). 

\subsection{Impact of Convolution Type}
\begin{figure}
    \centering
    \includegraphics[width=\linewidth]{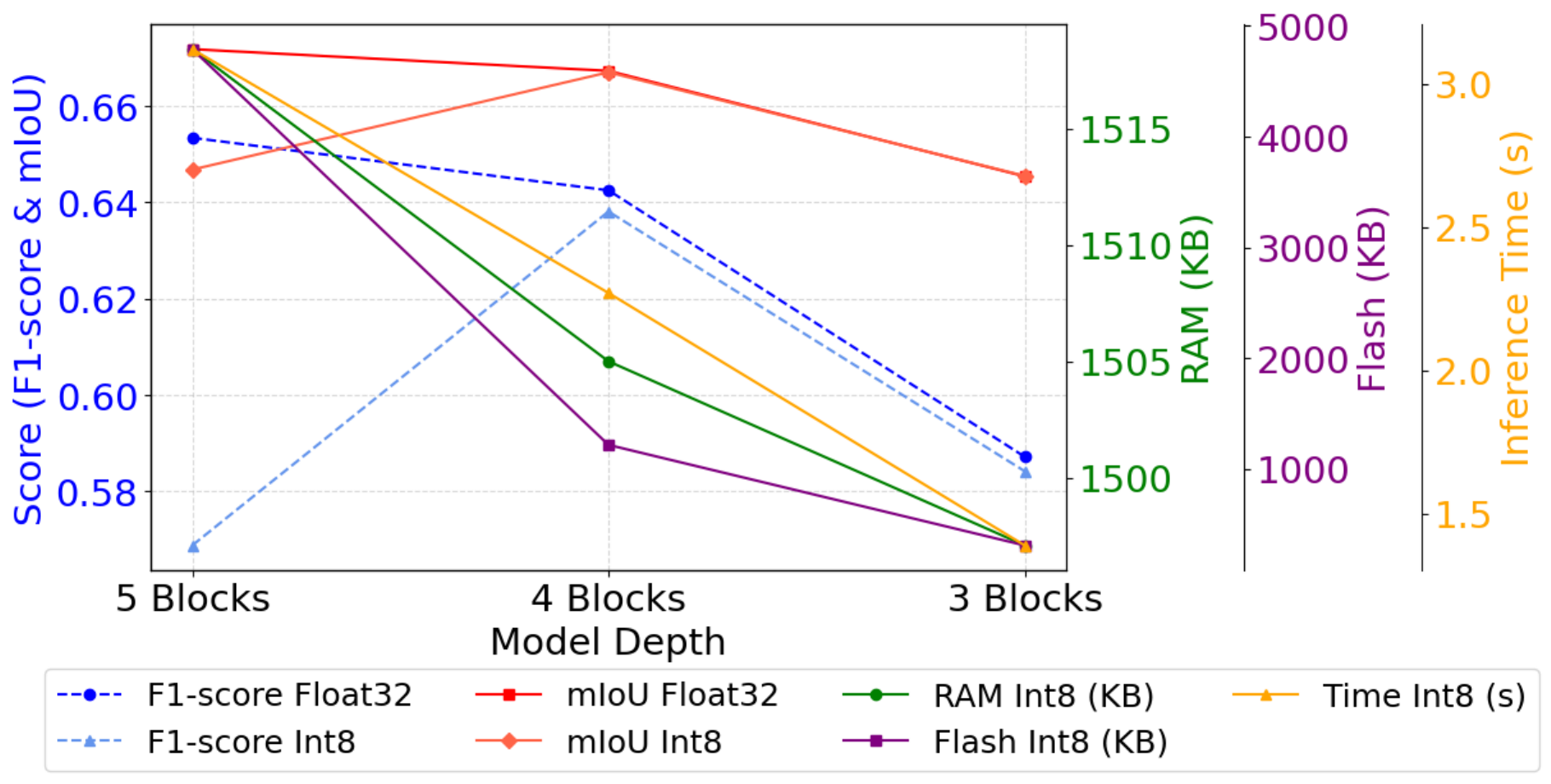}
    \caption{Average F1-score, mIoU, RAM, flash, and inference time vs model depth}
    \label{fig:modeldepth}
\end{figure}
\begin{figure}
    \centering
    \includegraphics[width=\linewidth]{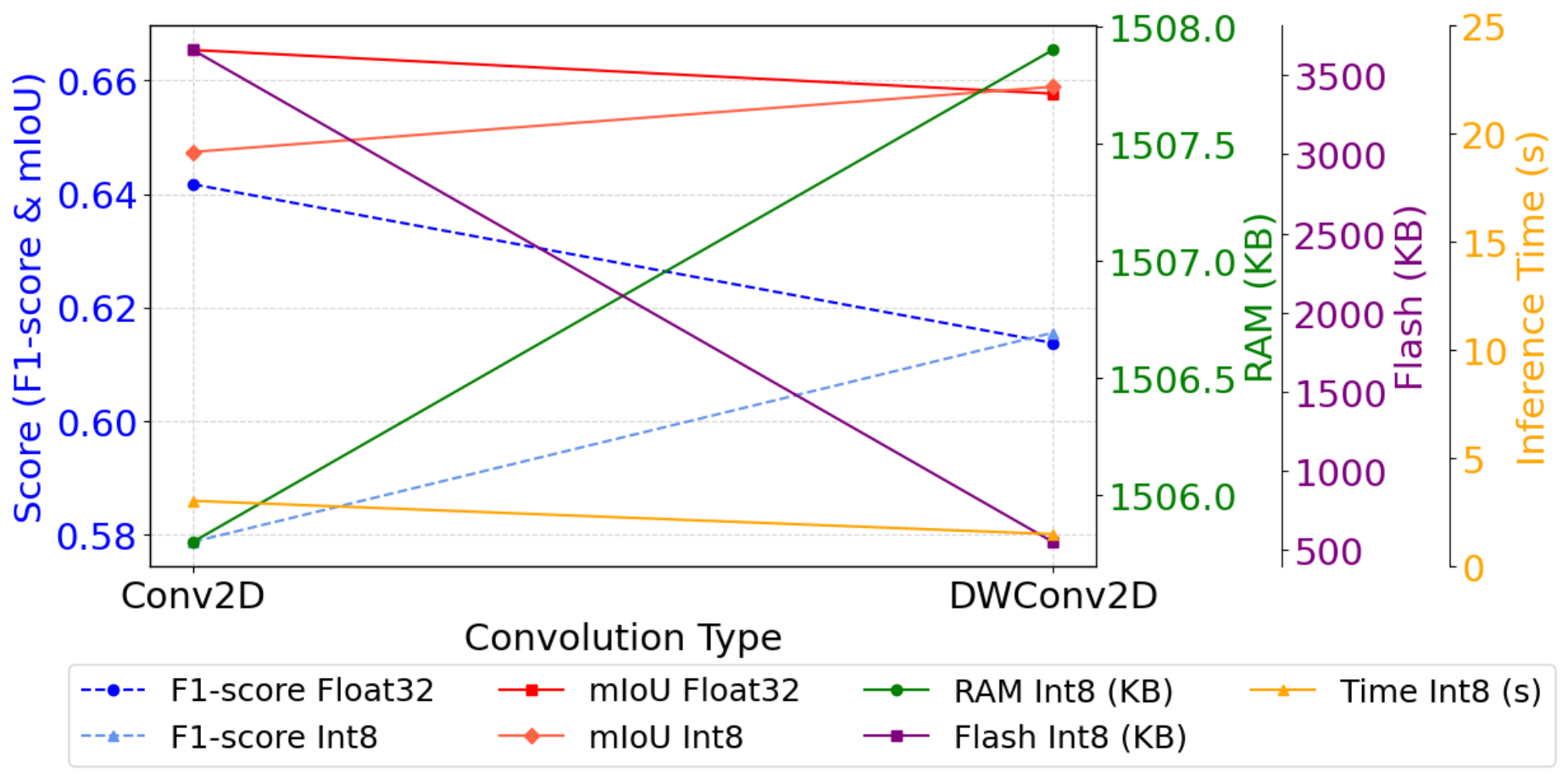}
    \caption{Average F1-score, mIoU, RAM, flash, and inference time vs convolution type}
    \label{fig:convtpye}
\end{figure}

This section examines the impact of varying the convolution type (comparing Conv2D and DWConv2D) on the segmentation performance and resource consumption of U-Net models configured with 5 ConvBlocks and x1 filter number shown in Fig. \ref{fig:convtpye}. The results presented here are averaged over all other parameters.

Under Float32 quantization, the Conv2D variant achieves an F1-score of 0.6417 and an mIoU of 0.6654, whereas the DWConv2D variant shows slightly reduced performance with an F1-score of 0.6137 and an mIoU of 0.6577. In contrast, for the Int8 quantized models, the DWConv2D configuration outperforms its Conv2D counterpart, yielding an F1-score of 0.6155 compared to 0.5786 and an mIoU of 0.6589 versus 0.6474. In terms of resource consumption, RAM usage remains comparable between the two configurations (approximately 1505.8 KB for Conv2D and 1507.9 KB for DWConv2D). However, the DWConv2D variant offers substantial benefits in Flash usage (reducing consumption from 3659.8 KB to 552.4 KB) and in inference time, which decreases from 3.034 s to 1.496 s.

\subsection{Overall Impact}
In summary, our comprehensive analysis indicates that an optimal trade-off between segmentation performance and resource efficiency can be achieved by judiciously selecting model configurations in each reduction strategy. Regarding filter number reduction, although full-scale configurations (x1 or x1/2) deliver the highest accuracy, they impose severe penalties in terms of RAM, Flash, and inference time. Conversely, overly aggressive reduction (x1/8 or x1/16) drastically diminishes performance. The intermediate x1/4 configuration thus emerges as the most promising candidate for balancing accuracy and resource savings. In the network depth reduction study, reducing the depth from 5 to 4 blocks results in only marginal declines in F1-score and mIoU while significantly reducing Flash consumption and inference latency compared to a 5-block design, making the 4-block configuration the preferable option. Finally, substituting standard Conv2D layers with DWConv2D yields substantial resource savings (most notably an 85\% reduction in Flash usage and a nearly 50\% decrease in inference time) with minimal impact on segmentation accuracy, or even slight improvements under Int8 quantization. Integrating these local optima, the overall best configuration is identified as a U-Net model with x1/4 filter numbers, a network depth of 4 blocks, and DWConv2D layers, which together offer a superior balance between segmentation performance and deployment efficiency for resource-constrained applications. Compared to the baseline U-Net (F1-score 0.618, mIoU 0.658, RAM 3,807.9 KB, flash 30,522.7 KB, not deployable), the best configuration improved the F1-score to 0.652 and mIoU to 0.676 while drastically cutting RAM to 952.5 KB and flash to 108 KB, ensuring deployability with an inference time of 1.23 s at 1.6605 J per inference.

\section{Conclusion and Future Work}

This study systematically explores model size reduction strategies for U-Net-based crack segmentation, tailored for TinyML deployments on a resource-constrained MCU. By investigating filter number and network depth reduction, and the use of Depthwise Separable Convolutions, we identified optimal configurations that balance segmentation accuracy and resource efficiency.

The results show that reducing convolution kernels and network depth significantly decreases RAM, Flash, and inference time consumption. The x1/4 filter number and 4-block network depth with DWConv2D configurations provide the best compromise, maintaining high segmentation performance while effectively reducing MCU resource usage. Comparing with one of the state-of-the-art crack segmentation RUCNet, this combination reduces the parameter number from 25.47M to 69.84K and still achieves 0.676 mIoU real-time crack segmentation on a low-power resource-constrained device. This study achieve a favorable balance between accuracy and deployment efficiency, making it particularly suitable for edge SHM systems. 

Future research will further explore ways to enhance model performance and plans to integrate energy harvesting modules, aiming to develop energy-autonomous crack monitoring sensor nodes for more efficient structural health monitoring.

\section*{Acknowledgement}
The authors would like to thank the financial support by the Knowledge Foundation under grant 20180170 (NIIT) and 20240029-H-02 (TransTech2Horizon).

\balance
\bibliographystyle{./IEEEtran}
\bibliography{IEEEsas2025}
\vspace{12pt}
\end{document}